\documentclass[10pt,twocolumn,letterpaper]{article}

\usepackage[pagenumbers]{cvpr} % To force page numbers, e.g. for an arXiv version

\usepackage{graphicx}
\usepackage{booktabs}
\usepackage{subfloat}
\usepackage{amsthm,amsmath,amssymb}
\usepackage{mathrsfs}
\usepackage{bm}
\usepackage{bbm}
\usepackage{multirow}
\usepackage[accsupp]{axessibility} % Improves PDF readability for those with disabilities.

\usepackage[pagebackref,breaklinks,colorlinks]{hyperref}

\usepackage[capitalize]{cleveref}
\crefname{section}{Sec.}{Secs.}
\Crefname{section}{Section}{Sections}
\Crefname{table}{Table}{Tables}
\crefname{table}{Tab.}{Tabs.}

\def\cvprPaperID{1779} % *** Enter the CVPR Paper ID here
\def\confName{CVPR}
\def\confYear{2022}

\begin{document}

\title{Trustworthy Long-Tailed Classification}

% \author{Bolian Li, Zongbo Han, Changqing Zhang\thanks{Corresponding author.}, \\
% Tianjin University, China \\
% {\tt\small \{libolian,zongbo,zhangchangqing\}@tju.edu.cn}
\author{
Bolian Li \\
Tianjin University \\
{\tt\small libolian@tju.edu.cn}
\and
Zongbo Han \\
Tianjin University \\
{\tt\small zongbo@tju.edu.cn}
\and
Haining Li\\
Xidian University\\
{\tt\small 18200100006@stu.xidian.edu.cn}
\and
Huazhu Fu\\
IHPC, A*STAR\\
{\tt\small hzfu@ieee.org}
\and
Changqing Zhang\thanks{Corresponding author.}\\
Tianjin University\\
{\tt\small zhangchangqing@tju.edu.cn}
}
\maketitle

%%%%%%%%% ABSTRACT
\begin{abstract}
Classification on long-tailed distributed data is a challenging problem, which suffers from serious class-imbalance and accordingly unpromising performance especially on tail classes. Recently, the ensembling based methods achieve the state-of-the-art performance and show great potential. However, there are two limitations for current methods. First, their predictions are not trustworthy for failure-sensitive applications. This is especially harmful for the tail classes where the wrong predictions is basically frequent. Second, they assign unified numbers of experts to all samples, which is redundant for easy samples with excessive computational cost. To address these issues, we propose a Trustworthy Long-tailed Classification (TLC) method to jointly conduct classification and uncertainty estimation to identify hard samples in a multi-expert framework. Our TLC obtains the evidence-based uncertainty (EvU) and evidence for each expert, and then combines these uncertainties and evidences under the Dempster-Shafer Evidence Theory (DST). Moreover, we propose a dynamic expert engagement to reduce the number of engaged experts for easy samples and achieve efficiency while maintaining promising performances. Finally, we conduct comprehensive experiments on the tasks of classification, tail detection, OOD detection and failure prediction. The experimental results show that the proposed TLC outperforms existing methods and is trustworthy with reliable uncertainty.
\end{abstract}

%%%%%%%%% BODY TEXT
\section{Introduction\label{sec:introduction}}
Data in real-world applications are usually long-tailed distributed over a series of categories~\cite{wang2020long,liu2019large,reed2001pareto,van2017devil,krishna2017visual,lin2014microsoft}. The frequencies of different categories vary a lot, with the head classes abundant in training samples, and the tail classes having only few training samples. Besides, there may also be new categories which models have not seen before~\cite{liu2019large}, exceeding the tail of long-tailed distribution and being termed as out-of-distribution (OOD) data~\cite{liang2018enhancing}. The long-tailed classification is very challenging since models need to handle the few-shot learning problem (and even with OOD data sometimes) for the tail classes, and the overall class-imbalance (models are trained on much more head samples than tail samples) would also deviate the models to focus extremely on the head classes~\cite{cao2019learning}. These problems cause the models to perform unpromisingly especially on the tail classes~\cite{he2009learning,buda2018systematic}.

Existing algorithms address long-tailed classification mainly by rebalancing the training of different classes to assign larger importance to tail samples~\cite{lin2017focal,cui2019class,wu2020distribution,cao2019learning}, transferring knowledge between the head and tail classes~\cite{liu2019large,zhu2020inflated}, ensembling statically sampled data groups~\cite{xiang2020learning,zhou2020bbn} (complementary ensembling), or ensembling individual classifiers in a multi-expert framework~\cite{wang2020long} (redundant ensembling). The redundant ensembling achieves the state-of-the-art performance mainly by reducing the model variance to obtain robust predictions~\cite{wang2020long}.
However, there are two major limitations for redundant ensembling methods. First, they are usually vulnerable to yielding unreliable prediction (i.e., over-confident prediction). This also prevents the ensembling methods from perceiving the wrong predictions and OOD samples, and is especially harmful for the tail classes where the predictions have averagely more errors than the head classes~\cite{he2009learning,buda2018systematic}. Consequently, their deployment in some failure-sensitive applications (e.g., disease diagnosis~\cite{ao2020application}, automatic driving~\cite{yasunobu2003auto} and robotics~\cite{davies2000review}) is limited.
Second, redundant ensembling usually assumes that all classifiers should be trained on all samples~\cite{wang2020long}, which is static and often induces excessive computational cost by uniformly assigning experts to all classes. The expert redundancy is severe especially on head classes, where competitive classification performance can be achieved with much fewer experts.

For these issues, we propose a novel Trustworthy Long-tailed Classification (TLC) method to jointly conduct classification and uncertainty estimation in a unified framework. First, we introduce the evidence and its associated uncertainty under the Dempster-Shafer Evidence Theory (DST)~\cite{dempster1967upper}. With the help of evidence-based uncertainty (EvU), our model can perceive hard samples in long-tailed classification, promoting the trustworthiness by detecting the tail and OOD samples, and identifying potentially wrong predictions. Second, we propose to combine the evidences from different experts with a uncertainty-based multi-expert fusion strategy under the Dempster's rule. We leverage the advantages of multiple experts to obtain accurate uncertainty and robust prediction. Moreover, we propose to reduce the number of engaged experts dynamically for the easy samples to jointly promote the efficiency while maintaining promising performances. For example, the actually needed number of experts for the head classes is less than that for tail classes (the head classes contain more easy samples). Therefore, we need to dynamically assign fewer experts in the training of head classes for efficiency. We achieve the dynamic expert engagement by incrementally adding experts when the previously added experts are all uncertain about their predictions. The main contributions are summarized as follows:
\begin{itemize}
    \item We introduce the evidence-based uncertainty (EvU) to promote the trustworthiness of long-tailed classification. To the best of our knowledge, the proposed TLC is the first work asserting trustworthiness in long-tailed classification. 
    \item We propose a multi-expert fusion strategy based on the uncertainty of each expert under the Dempster-Shafer Evidence Theory (DST), which promotes the classification performance and trustworthiness by reliably perceiving hard samples.
    \item We achieve efficiency in training multiple experts by dynamically reducing the engaged experts with uncertainty, and obtain promising performances meanwhile.
    \item We conduct experiments on classification, tail \& OOD sample detection and failure prediction, and evaluate the results with diverse metrics, which validates that the proposed TLC outperforms existing methods in the above tasks and is trustworthy with reliable uncertainty. The code\footnote{\url{https://github.com/lblaoke/TLC}} is publicly available.
\end{itemize}

\begin{figure*}[ht]
    \centering
    \subfloat[Training\label{subfig:train}]{\includegraphics[width=8.5094cm]{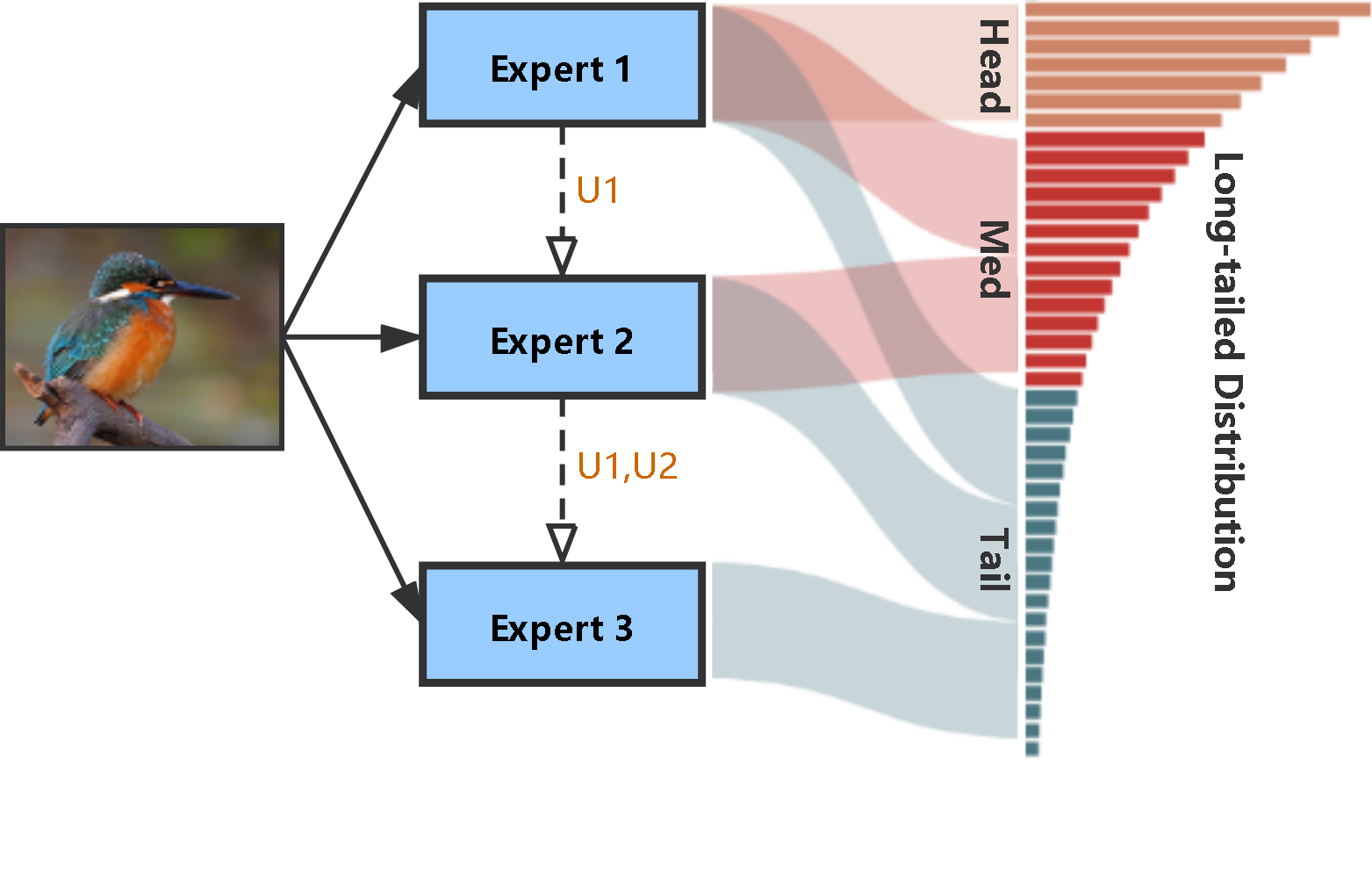}}
    \subfloat[Testing\label{subfig:test}]{\includegraphics[width=8.8906cm]{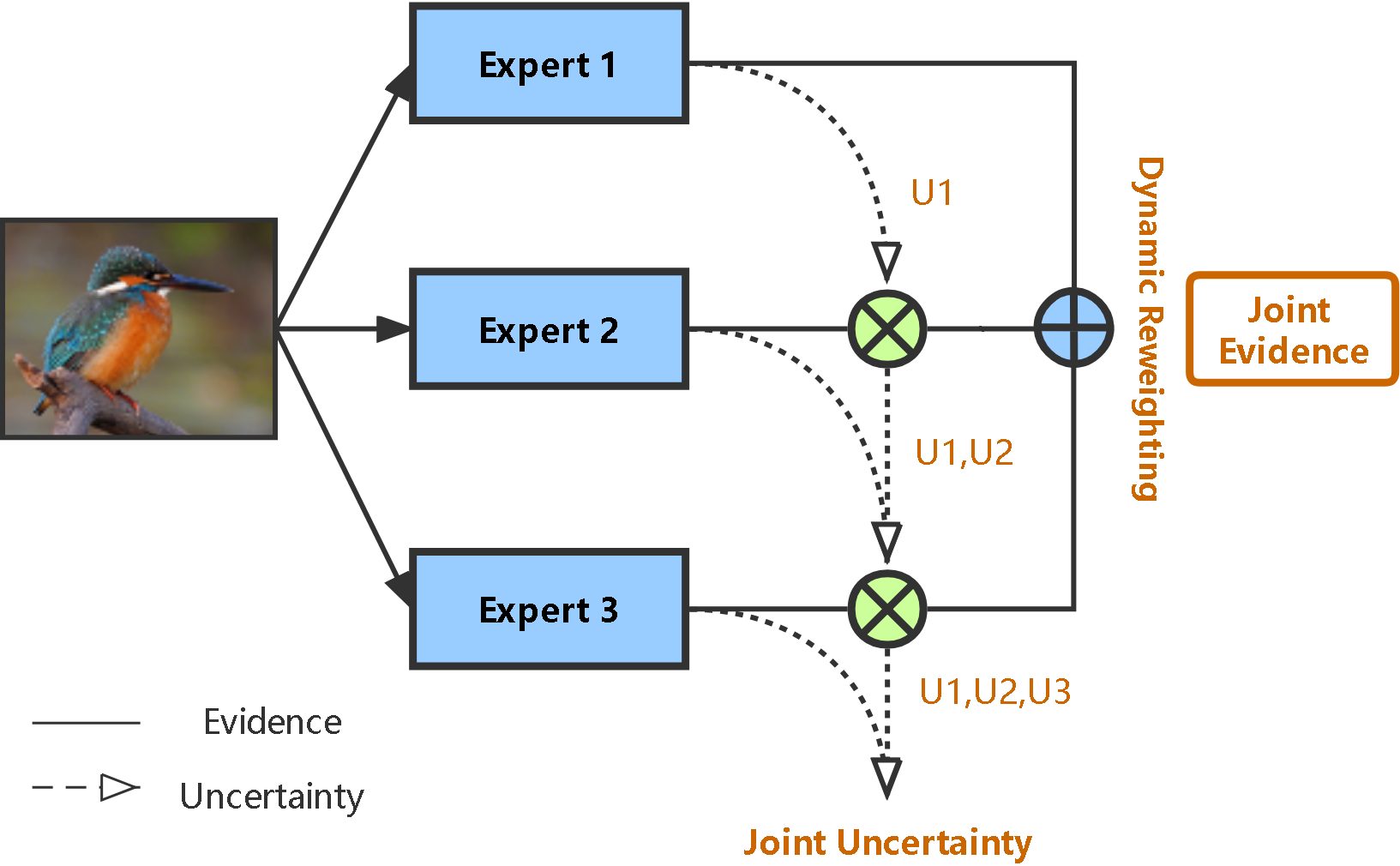}}
    \caption{Overview of the proposed Trustworthy Long-Tailed Classification (TLC). U1, U2 and U3 are the uncertainties of expert $1$, $2$ and $3$ respectively. In training (a), we provide an example of collaborating in different class groups for multiple experts. TLC dynamically assigns averagely more experts to the samples in tail classes than those in head classes. This assignment is achieved automatically by identifying hard samples with uncertainty. In testing (b), the joint uncertainty is formed with the Dempster's rule, and the joint evidence is obtained by uncertainty-based dynamic reweighting.}
    \label{fig:TLC}
\end{figure*}

\section{Related Work\label{sec:relate}}
\noindent\textbf{Long-tailed classification.}
Traditional long-tailed classification methods include under-sampling~\cite{liu2008exploratory}, over-sampling~\cite{han2005borderline} and data augmentation~\cite{liu2020deep,chu2020feature,kim2020m2m}. Re-balancing methods~\cite{lin2017focal,cui2019class,wu2020distribution,cao2019learning,menon2020long} focus more on the tail classes. OLTR~\cite{liu2019large} and inflated memory~\cite{zhu2020inflated} transfer the knowledge between different class regions. BBN~\cite{zhou2020bbn} learns the head and tail patterns separately. LFME~\cite{xiang2020learning} distills multiple teacher models respectively for class regions. RIDE~\cite{wang2020long} and ACE~\cite{cai2021ace} ensembles multiple experts to obtain robust predictions. TDE~\cite{tang2020long} adopts casual inference to eliminate the biases of tail classes. These methods do not fully explore the uncertainty for perceiving hard samples in the predictions.

\noindent\textbf{Uncertainty estimation.}
Traditional uncertainty estimation algorithms are discussed in~\cite{blatz2004confidence, abdar2021review}. BNN~\cite{blundell2015weight} models uncertainty by replacing the deterministic parameters with distributions. MC Dropout~\cite{gal2016dropout} approximates BNN with dropout. MCP~\cite{hendrycks2016baseline} obtains uncertainty from the softmax distribution. TCP~\cite{corbiere2019addressing} learns an extra module to yield uncertainty. EDL~\cite{csensoy2018evidential} models uncertainty under the subjective logic. Ensembling methods like \cite{lakshminarayanan2017simple} obtains uncertainty from the diverse predictions. DUQ~\cite{van2020uncertainty} estimates the RBF distances as uncertainty. GP~\cite{damianou2013deep} models uncertainty as the similarity between samples using non-parametric kernel function.

\noindent\textbf{Evidence theory.}
The Dempster-Shafer Evidence Theory (DST) was first proposed by \cite{dempster1967upper}. It was later generalized as a framework to model the epistemic uncertainty~\cite{shafer1976mathematical}. The DST formulates the Bayesian inference with the subjective logic~\cite{dempster1968generalization}. The DST allows the beliefs from different sources to be combined into a joint belief~\cite{sentz2002combination,josang2012interpretation}.

\section{Preliminaries\label{sec:pre}}
A long-tailed dataset consists of an imbalanced training set and a balanced test set. Formally, we define an input $\bm{x}_i\in\mathbbm{R}^d$, its corresponding label $y_i\in\{1,2,\cdots, K\}$, and the class-conditional distribution $p(\bm{x}|y)$. For the training set, the following relationships holds:
\begin{equation}
    \left\{
    \begin{aligned}
        &\int p(\bm{x}~|~y=k_1)d\bm{x}\geq \int p(\bm{x}~|~y=k_2)d\bm{x},~\forall~k_1\leq k_2 \\
        &\lim_{k\rightarrow\infty}\int p(\bm{x}~|~y=k)d\bm{x}=0
    \end{aligned}
    \right.,
\end{equation}
indicating that the class volumes decay successively with the ascending class indexes and finally approach zero in the last few classes. The classes can be separated into the head, medium and tail regions based on different numbers of samples. For the test set, following the setting in most exising works on long-tailed problem~\cite{cao2019learning,liu2019large,wang2020long,cui2019class}, the class frequencies are equal for the fairness among categories:
\begin{equation}
    \int p(\bm{x}~|~y=k_1)d\bm{x}=\int p(\bm{x}~|~y=k_2)d\bm{x},~\forall~k_1,k_2.
\end{equation}
Additionally, for datasets that are not naturally long-tailed, a widely used transformation is to sample a subset with a specific decay distribution (e.g., exponential distribution~\cite{cui2019class} and Pareto distribution~\cite{liu2019large}).
% \begin{equation}
%     p'(\bm{x}~|~y=k)=P_{decay}\{y=k\}\cdot p(\bm{x}~|~y=k),
% \end{equation}
% which randomly removes samples from class $k$ with probability $1-P_{decay}\{y=k\}$.

\section{Method\label{sec:method}}
In this section, we introduce how to estimate uncertainty with Dempster-Shafer Evidence Theory in Sec.~\ref{ssec:eu}, propose to form joint uncertainty and joint evidence with the Dempster's rule in Sec.~\ref{ssec:combine}, and show the training process with dynamic expert engagement in Sec.~\ref{ssec:learn}.

\subsection{Estimating Evidence-based Uncertainty\label{ssec:eu}}
In long-tailed classification, perceiving hard samples with uncertainty can reduce the cost of trusting wrong predictions, which is especially important in tail classes with few training samples. However, existing methods suffer from over-confidence ~\cite{moon2020confidence,van2020uncertainty} or excessive computational cost~\cite{blundell2015weight,gal2016dropout,corbiere2019addressing}. Therefore, for trustworthy long-tailed classification, we introduce the evidence-based uncertainty (EvU) under the Dempster-Shafer Evidence Theory (DST) to promote trustworthiness and efficiency simultaneously.

The DST is a generalization of the Bayesian theory of subjective probability~\cite{dempster1968generalization}. While based on DST, subjective logic (SL) explicitly takes epistemic uncertainty and source trust into account \cite{audun2018subjective}. The DST assigns \textbf{belief masses} to the possible sets of class labels for a prediction, measuring the chances to find the true class labels in these sets~\cite{csensoy2018evidential}. When a belief mass is assigned to all class labels, these classes are equally likely. Therefore, such belief mass can represent the \textbf{uncertainty} of the entire prediction~\cite{audun2018subjective}. Formally, the subjective logic defines the belief assignment over a Dirichlet distribution~\cite{kotz2004continuous}:
\begin{equation}
    D(\bm{p}~|~\bm{\alpha})=\left\{
    \begin{aligned}
        &\frac{1}{B(\bm{\alpha})}\prod\nolimits_{k=1}^Kp_k^{\alpha_k-1}~\text{for}~\bm{p}\in\mathcal{S}_K \\
        &0~~~~~~~~~~~~~~~~~~~~~~~~~~~~~~~~\text{otherwise}
    \end{aligned}
    \right.,
    \label{eq:dirichlet}
\end{equation}
where $\bm{\alpha}$ are parameters of the distribution, $B(\cdot)$ is the beta function, and $\mathcal{S}_K=\{\bm{p}|\sum_{k=1}^Kp_k=1~\text{and}~0\leq p_k\leq1,\forall k\}$ is the $K$-dimensional unit simplex. The uncertainty and belief masses are determined by the parameters as:
\begin{equation}
    u=\frac{K}{S}~~\text{and}~~b_k=\frac{\alpha_k-1}{S},\label{eq:ub}
\end{equation}
where $\alpha_k$ is the Dirichlet parameter for the $k$-th class and $S=\sum_{k=1}^K\alpha_k$ is the Dirichlet strength. In this way, the uncertainty is less likely to suffer from over-confidence since it avoids merely regarding the probability of labeled class as uncertainty, and takes into account the comparative values against other classes~\cite{csensoy2018evidential}.

In DST, the \textbf{evidence} is a measure of the support derived from data in favor of a sample to fall into a certain class~\cite{csensoy2018evidential}. The evidence of each class $\bm{e}=[e_1,e_2,\cdots,e_K]$ can be obtained directly from the output of neural networks by replacing the softmax layer with a non-negative activation function. Then, the parameters of Dirichlet distribution in Eq.~\ref{eq:dirichlet} can be computed by using $\alpha_k=e_k+1$, and therefore, the uncertainty and belief masses can be quantified with Eq.~\ref{eq:ub}. Moreover, it is obvious that the total amount of uncertainty and belief masses is a constant, i.e.,  $u+\sum_{k=1}^Kb_k=1$ (implying a $K+1$-dimensional unit simplex). When the evidences on all classes are insufficient for a prediction, the belief masses assigned to all classes will also be low, and meanwhile, the uncertainty for this output will be high to indicate a high probability of erroneous prediction.\footnote{It is discussed in the supplementary material.\label{footnote}}

The strength of EvU (evidence-based uncertainty) lies in its modeling based on the Dirichlet distribution, which parameterizes the density of belief assignments directly from the outputs of neural networks. EvU models the uncertainty and high-order probabilities for a prediction~\cite{han2020trusted}. Further, EvU also theoretically avoids the over-confident problem (common in traditional uncertainty estimation algorithms~\cite{hendrycks2016baseline}) by obtaining the uncertainty from the overall belief masses. It is noteworthy that the EvU can be obtained directly with Eq.~\ref{eq:ub}, which is efficient and reasonable.

\subsection{Combining Experts with Dempster’s Rule\label{ssec:combine}}
We employ a multi-expert framework with each expert guided by the DST introduced in Sec.~\ref{ssec:eu}. \cite{wang2020long} shows that integrating multiple classifiers will reduce the model variance, which is beneficial to the robustness of long-tailed classification. On top of ensembling, we combine the uncertainties and evidences of multiple experts under the Dempster's rule (shown in Fig.~\ref{subfig:test}).

\noindent\textbf{Combining uncertainties.}
We combine the uncertainties of multiple experts in an incremental fashion (e.g., first combine expert $1$ and $2$, and then add on expert $3$). First, we formalize the pair-wise Dempster's combination rule as:
\begin{equation}
    u^1\oplus u^2=\frac{1}{1-C}u^1u^2,
\end{equation}
where $C=\sum_{i\neq j}b_i^1b_j^2$ is the conflict factor. When two experts agree on most of belief masses (i.e., $C$ is small), the combined uncertainty will be relatively low.\textsuperscript{\ref{footnote}} Second, we apply the combination rule to sequentially combine multiple experts, and the final combination is induced as:
\begin{equation}
    u=u^1\oplus u^2\oplus......\oplus u^M=\frac{\prod_{m=1}^Mu^m}{\prod_{m=1}^M(1-C^m)},\label{eq:jointu}
\end{equation}
where $C^m=\sum_{i\neq j}b_i^mb_j^{m-1}$ is the conflict factor between two consecutive experts and $C^1=0$ (the first expert do not have former result to compare with). The uncertainty combination considers both independent uncertainty from each expert and the agreement of beliefs between different experts.

\noindent\textbf{Combining evidences.}
We dynamically reduce the engaged experts on easy samples in training stage (detailed discussion in Sec.~\ref{ssec:learn}). Therefore, at test stage, the engaged experts for easy samples should also be limited. For example, for the head classes, we primarily consider the evidence of the first few experts, while for the tail classes, the evidences of all experts should be considered. First, we define the \textbf{prefix weights} of each expert with the following rules:
\begin{equation}
\left\{
\begin{aligned}
    &w^1=1; \\
    &w^2=u^1; \\
    &w^{m+1}=w^m\oplus u^m=\frac{1}{1-C^m}w^mu^m, \\
    &~~~~~~~~~~~~~~~~~~~~~~~~~~~~~~~~\text{for}~m=2,3,\cdots,M-1.\label{eq:prefix}
\end{aligned}
\right.
\end{equation}
The prefix weight $w^m$ is a measure of the overall uncertainty from experts previous to expert $m$. It accords with the combining process of the joint uncertainty in Eq.~\ref{eq:jointu}, and regards the intermediate combination results as weights. When the experts previous to expert $m$ are already certain about their evidences (indicating that it is an easy sample), the prefix weight $w^m$ will be low (indicating that $\bm{e}^m$ is not obliged to consider). Second, we apply the prefix weights to combine the evidences at inference time:
\begin{equation}
    \bm{e}=\frac{\sum_{m=1}^M\exp\{w^m/\eta\}\cdot \bm{e}^m}{\sum_{m=1}^M\exp\{w^m/\eta\}},
\end{equation}
where exponentiation is adopted for non-maximum suppression~\cite{neubeck2006efficient}, and $\eta$ is a temperature factor which adjusts the sensitivity of the prefix weights. We also apply the prefix weights for training efficiency in Sec.~\ref{ssec:learn}.

\subsection{Learning Experts with Dynamic Engagement\label{ssec:learn}}
In our multi-expert framework, each expert can capture the evidence from input to induce a classification opinion~\cite{kiela2018efficient}. We propose to jointly learn experts under the subjective logic, while dynamically reducing the number of engaged experts for easy samples.

\noindent\textbf{Learning single expert.} For a single expert, we formulate the objective with the Type \uppercase\expandafter{\romannumeral2} Maximum Likelihood (Empirical Bayes)~\cite{jamil2012selection}. First, we obtain the evidence $\bm{e}_i$ and convert class label $y_i$ into a one-hot vector $\bm{y}_i$. Second, we treat an adjusted Dirichlet distribution $\widetilde{D}(\bm{p}_i|\bm{e}_i)$ as the prior of multinomial likelihood $P(\bm{y}_i|\bm{p}_i)$ (the classification opinion) and then compute the negative logarithm of the marginal likelihood:
\begin{equation}
\begin{aligned}
    \mathcal{L}&=-\log\left[\int\prod_{k=1}^Kp_{ik}^{y_{ik}}\frac{1}{B(\bm{e}_i)}\prod_{k=1}^Kp_{ik}^{e_{ik}-1}d\bm{p}_i\right] \\
    &=\sum_{k=1}^Ky_{ik}\left[\log\left(\sum\nolimits_{k=1}^Ke_{ik}\right)-\log(e_{ik})\right].
    \label{eq:l}
\end{aligned}
\end{equation}
However, the objective in Eq.~\ref{eq:l} only ensures that the correct class will be assigned with more evidence than other classes, while there is no support for the low overall evidences on the incorrect classes. In another word, the uncertainty may be unreasonably low due to the high overall evidences. We address this problem by introducing the following Kullback-Leibler divergence~\cite{csensoy2018evidential}:
\begin{equation}
\begin{aligned}
    \mathcal{L}_{kl}=&KL(D(\bm{p}_i~|~\widetilde{\bm{\alpha}}_i)~||~D(\bm{p}_i~|~\bm{1})) \\
    =&\log\frac{\Gamma(\widetilde{S}_i)}{\Gamma(K)\prod_{k=1}^K\Gamma(\widetilde{\alpha}_{ik})} \\
    &~+\sum_{k=1}^K(\widetilde{\alpha}_{ik}-1)\left[\psi(\widetilde{\alpha}_{ik})-\psi\left(\widetilde{S}_i\right)\right],
\end{aligned}
\end{equation}
where $\widetilde{\bm{\alpha}}_i=\bm{1}+(\bm{1}-\bm{y}_i)\odot\bm{e}_i$ is the adjusted Dirichlet parameters, $\widetilde{S}_i=\sum_{k=1}^K\widetilde{\alpha}_k$ is the adjusted Dirichlet strength, $\Gamma(\cdot)$ is the gamma function, and $\psi(\cdot)$ is the digamma function. This KL divergence regulates the evidences of incorrect classes to 0 by minimizing the distance between the adjusted distribution and the target distribution, and thus can avoid the uncertainty to be unreasonably low. Finally, the objective for a single expert is
\begin{equation}
    \mathcal{L}_{single}=\mathcal{L}+\lambda_{kl}(t)\mathcal{L}_{kl},
\end{equation}
where $\lambda_{kl}(t)=min\{1,t/T\}$ is the annealing factor ($t$ is the current epoch). We gradually increase the the KL divergence to prevent the expert to learn a flat uniform evidence in the early stage of training.

\noindent\textbf{Learning multiple experts dynamically.}
We argue that there is no necessity in learning easy samples (usually in head classes) with all experts, since using fewer experts can also achieve competitive performances for these samples. We show the experimental support for this in Sec.~\ref{ssec:visual}. To this end, we propose to apply the prefix weights of all experts (a measure of joint uncertainty from a group of experts defined in Eq.~\ref{eq:prefix}) to dynamically remove the losses on easy samples. For example, if experts $1, 2, \cdots, m-1$ are all certain about a sample (i.e., $w^m\leq\tau$), the loss of expert $m$ on this sample will be removed. Therefore, the overall expert engagement should be in a descending pattern. For example, the first expert is responsible for all classes, the second expert for the classes except the head classes, and the last expert only focuses on the tail classes (as shown in Fig.~\ref{subfig:train}).

Additionally, to enhance the diversity of experts, we form the output distribution $P(\bm{p}_i|\bm{\alpha}_i^m)$ with the normalized Dirichlet parameters: $P(p_{ik}|\bm{\alpha}_i^m)=\alpha_{ik}^m/S_i^m$, and push different experts apart by the following KL divergence:
\begin{equation}
    \mathcal{L}_{div}=-\frac{1}{M}\sum\nolimits_{m=1}^MKL(P(\bm{p}_i~|~\bm{\alpha}_i^m)~||~P(\bm{p}_i~|~\overline{\bm{\alpha}}_i)),
\end{equation}
where $\overline{\bm{\alpha}}_i=\sum_{m=1}^M\bm{\alpha}_i^m/M$ are the averaged Dirichlet parameters.

Finally, the joint objective to learn evidences in the multi-expert framework is given by adding up the objective of each expert:
\begin{equation}
    \mathcal{L}=\sum_{i=1}^N\sum_{m=1}^M\mathbbm{1}\{w_i^m>\tau\}\mathcal{L}_{single}+\lambda_{div}\mathcal{L}_{div}.
\end{equation}
The overall training process is summarized in the supplementary material.

\section{Experiments\label{sec:experiment}}
In this section, we conduct experiments to answer the following questions:
\begin{itemize}
    \item \textbf{Q1 (Effectiveness):} Does the proposed TLC outperform the state-of-the-art methods in long-tailed classification? (Sec.~\ref{ssec:quant})
    \item \textbf{Q2 (Trustworthiness \uppercase\expandafter{\romannumeral1}):} How to validate the trustworthiness of TLC, and is the estimated uncertainty reliable? (Sec.~\ref{ssec:quant} and Sec.~\ref{ssec:ablation})
    \item \textbf{Q3 (Efficiency \uppercase\expandafter{\romannumeral1}):} Why is it reasonable to reduce the engaged experts for easy samples? (Sec.~\ref{ssec:visual})
    \item \textbf{Q4 (Trustworthiness \uppercase\expandafter{\romannumeral2}):} Is the estimated uncertainty good at discerning the head, medium and tail classes? (Sec.~\ref{ssec:visual})
    \item \textbf{Q5 (Efficiency \uppercase\expandafter{\romannumeral2}):} Does the actual expert engagement accord with our expectation in Sec.~\ref{ssec:learn}? (Sec.~\ref{ssec:visual})
\end{itemize}
Specifically, we show the configurations in Sec.~\ref{ssec:setup}, quantitative and qualitative results in Sec.~\ref{ssec:quant} and Sec.~\ref{ssec:visual} respectively, and ablation studies in Sec.~\ref{ssec:ablation}.

\subsection{Experimental Setup\label{ssec:setup}}
\noindent\textbf{Tasks.}
Along with classification, to show the effect of uncertainty in long-tailed problem, we conduct the following tasks: tail detection, out-of-distribution (OOD) detection and failure prediction~\cite{hendrycks2016baseline}. These tasks all use the estimated uncertainty for binary classification. Specifically, in tail and OOD detection, uncertainties are used to distinguish the tail/OOD samples from others, and in failure prediction, uncertainties are used to distinguish between incorrect and erroneous predictions. The metrics for evaluation are similar to those used in binary classification and confidence calibration (e.g., AUC~\cite{mcclish1989analyzing}, FPR-95~\cite{liang2017principled} and ECE~\cite{naeini2015obtaining}).

\noindent\textbf{Datasets.} We use three long-tailed datasets (CIFAR-10-LT, CIFAR-100-LT and ImageNet-LT) and three balanced OOD datasets (SVHN~\cite{netzer2011reading}, ImageNet-open and Places-open). CIFAR-10-LT and CIFAR-100-LT~\cite{cui2019class} are sampled from the original CIFAR~\cite{krizhevsky2009learning} dataset over exponential distributions ~\cite{cui2019class}. ImageNet-LT~\cite{liu2019large} is sampled from the ImageNet-2012~\cite{deng2009imagenet} dataset over Pareto distributions with the power value $\alpha=6$. It contains 115.8K images in 1,000 classes. ImageNet-open is the additional classes of images in the ImageNet-2010 dataset~\cite{liu2019large}. Places-open~\cite{liu2019large} is the test images from the Places-Extra69 dataset~\cite{zhou2017places}.

\noindent\textbf{Compared methods.} We compare the proposed TLC with re-balancing methods including Focal Loss~\cite{lin2017focal}, LDAM-DRW~\cite{cao2019learning}, $\tau$-norm and cRT~\cite{kang2019decoupling}, knowledge transfer method OLTR~\cite{liu2019large}, and ensemble learning method RIDE~\cite{wang2020long}. We also compare the evidence-based uncertainty with other widely used uncertainty estimation algorithms including the Maximal Class Probability (MCP)~\cite{hendrycks2016baseline}, Gaussian Process (GP)~\cite{damianou2013deep}, and Monte Carlo Dropout (MCD)~\cite{gal2016dropout}.

\subsection{Quantitative Evaluation\label{ssec:quant}}
\begin{table*}[ht]\small%
\renewcommand{\arraystretch}{0.84}
\centering
\caption{Performance comparison on long-tailed classification in terms of ACC (in percentage).}
\label{tab:classification}
\begin{tabular}{c|c|ccccc}
\hline
Dataset                       & Method         & All                   & Region                & Head                  & Med                   & Tail                  \\ \hline\hline
\multirow{9}{*}{CIFAR-10-LT}  & Focal Loss     & 68.6$\pm$0.2          & 73.2$\pm$0.4          & 84.8$\pm$0.2          & 67.9$\pm$0.9          & 49.1$\pm$0.8          \\
                              & OLTR           & 78.7$\pm$0.6          & 80.5$\pm$0.3          & 86.1$\pm$0.1          & 77.5$\pm$0.6          & 69.8$\pm$1.7          \\
                              & LDAM-DRW       & 78.4$\pm$1.0          & 82.5$\pm$0.4          & \textbf{89.6$\pm$0.1} & 74.0$\pm$1.5          & 72.4$\pm$2.0          \\
                              & $\tau$-norm    & 79.6$\pm$1.0          & 83.5$\pm$0.4          & 87.7$\pm$0.2          & 76.2$\pm$1.6          & 73.6$\pm$1.4          \\
                              & cRT            & 79.2$\pm$0.3          & 83.0$\pm$0.4          & 87.1$\pm$0.1          & 77.3$\pm$0.8          & 71.5$\pm$0.9          \\
                              & RIDE           & 80.2$\pm$0.3          & 83.4$\pm$0.2          & 87.4$\pm$0.1          & 77.2$\pm$0.7          & 75.0$\pm$0.5          \\ \cline{2-7} 
                              & TLC(2 experts) & 80.3$\pm$0.4          & 84.2$\pm$0.3          & 86.0$\pm$0.1          & 77.8$\pm$0.5          & 75.4$\pm$0.8          \\
                              & TLC(3 experts) & 80.3$\pm$0.4          & 84.2$\pm$0.3          & 85.9$\pm$0.1          & 77.2$\pm$0.8          & \textbf{75.9$\pm$0.6} \\
                              & TLC(4 experts) & \textbf{80.4$\pm$0.2} & \textbf{84.4$\pm$0.2} & 85.7$\pm$0.1          & \textbf{78.1$\pm$0.5} & 75.6$\pm$0.5          \\ \hline\hline
\multirow{9}{*}{CIFAR-100-LT} & Focal Loss     & 42.3$\pm$1.3          & 55.4$\pm$0.4          & 70.3$\pm$1.7          & 40.7$\pm$1.6          & 15.9$\pm$1.9          \\
                              & OLTR           & 43.4$\pm$0.8          & 59.9$\pm$0.2          & 64.6$\pm$2.0          & 44.8$\pm$1.5          & 20.9$\pm$2.4          \\
                              & LDAM-DRW       & 44.4$\pm$1.2          & 61.4$\pm$0.2          & 64.8$\pm$1.5          & 43.8$\pm$1.3          & 24.6$\pm$1.8          \\
                              & $\tau$-norm    & 45.4$\pm$1.2          & 62.3$\pm$0.6          & 68.0$\pm$1.6          & 47.2$\pm$1.4          & 21.0$\pm$2.0          \\
                              & cRT            & 45.6$\pm$0.3          & 62.3$\pm$0.5          & 67.8$\pm$2.4          & 47.1$\pm$2.1          & 21.8$\pm$1.6          \\
                              & RIDE           & 48.3$\pm$0.5          & 62.8$\pm$0.1          & 68.8$\pm$1.2          & \textbf{49.0$\pm$0.7} & 27.1$\pm$1.4          \\ \cline{2-7} 
                              & TLC(2 experts) & 47.2$\pm$0.7          & 62.8$\pm$0.3          & 69.4$\pm$1.2          & 46.6$\pm$1.0          & 25.7$\pm$1.5          \\
                              & TLC(3 experts) & 49.0$\pm$0.4          & 64.0$\pm$0.2          & 70.9$\pm$0.8          & 47.9$\pm$0.9          & 28.1$\pm$1.3          \\
                              & TLC(4 experts) & \textbf{49.8$\pm$0.8} & \textbf{64.5$\pm$0.2} & \textbf{71.1$\pm$1.0} & 48.4$\pm$1.1          & \textbf{29.7$\pm$1.6} \\ \hline\hline
\multirow{9}{*}{ImageNet-LT}  & Focal Loss     & 45.6$\pm$2.1          & 67.0$\pm$0.6          & 69.2$\pm$3.2          & 41.5$\pm$2.7          & 26.1$\pm$3.1          \\
                              & OLTR           & 50.7$\pm$1.2          & 68.0$\pm$0.5          & 67.8$\pm$1.9          & 53.3$\pm$1.8          & 31.0$\pm$2.4          \\
                              & LDAM-DRW       & 49.8$\pm$0.7          & 66.9$\pm$0.5          & 63.3$\pm$2.1          & 50.2$\pm$2.2          & 36.0$\pm$1.3          \\
                              & $\tau$-norm    & 47.9$\pm$1.2          & 67.8$\pm$0.3          & 60.3$\pm$1.8          & 50.6$\pm$1.3          & 33.0$\pm$1.8          \\
                              & cRT            & 48.4$\pm$1.3          & 67.5$\pm$0.5          & 64.4$\pm$2.4          & 50.5$\pm$1.4          & 30.3$\pm$1.8          \\
                              & RIDE           & 54.6$\pm$0.9          & 68.4$\pm$0.3          & \textbf{70.6$\pm$1.3} & 54.8$\pm$0.9          & 38.3$\pm$1.4          \\ \cline{2-7} 
                              & TLC(2 experts) & 54.1$\pm$0.6          & 68.4$\pm$0.3          & 68.7$\pm$1.2          & 55.4$\pm$1.2          & 38.3$\pm$1.4          \\
                              & TLC(3 experts) & 54.6$\pm$0.5          & 69.1$\pm$0.3          & 69.3$\pm$1.2          & \textbf{56.7$\pm$0.8} & 37.9$\pm$1.8          \\
                              & TLC(4 experts) & \textbf{55.1$\pm$0.7} & \textbf{69.9$\pm$0.2} & 68.9$\pm$1.2          & 55.7$\pm$1.5          & \textbf{40.8$\pm$0.8} \\ \hline
\end{tabular}
\end{table*}

\noindent\textbf{Classification (Q1).}
We evaluate the performances on classification with diverse metrics. Along with the Top-1 accuracy, we also report the \textbf{regional accuracy} which computes the frequency of predictions falling into the correct class region (e.g., whether tail samples are classified into tail classes\footnote{When they fail, they are still likely to be trusted due to the lower averaged uncertainty of head classes, but even if they fall into other tail classes erroneously, the model is still uncertain about them, and thus reduce the potential threat.}).
Higher regional accuracy implies better trustworthiness for long-tailed classification.
The evaluation results are listed in Table.~\ref{tab:classification}. We run each experiment five times to report the average ACC and standard deviation. The proposed TLC outperforms the compared methods on all datasets, and improves the regional and tail ACC significantly.

\begin{table*}[ht]\small%
\renewcommand{\arraystretch}{0.84}
\setlength\tabcolsep{3.9pt}
\centering
\caption{Performances comparison on tail detection and OOD detection in terms of AUC (in percentage).}
\label{tab:ood}
\begin{tabular}{c|cccc|cccc|cc}
\hline
Training       & \multicolumn{4}{c|}{CIFAR-10-LT}                                                                                                                  & \multicolumn{4}{c|}{CIFAR-100-LT}                                                                                                                 & \multicolumn{2}{c}{ImageNet-LT}                                          \\ \hline
Testing        & Tail          & SVHN          & ImageNet-open & Places-open & Tail          & SVHN          & ImageNet-open & Places-open & Tail          & ImageNet-open \\ \hline
Focal Loss     & 36.3          & 64.0          & 70.0                                                     & 70.6                                                   & 35.4          & 54.0          & 53.5                                                     & 53.2                                                   & 26.8          & 43.1                                                     \\
OLTR           & 55.9          & 55.9          & 78.2                                                     & 77.1                                                   & 37.2          & 53.7          & 54.1                                                     & 52.8                                                   & 27.5          & 42.1                                                     \\
LDAM-DRW       & 54.9          & 55.8          & 78.2                                                     & 76.5                                                   & 36.9          & \textbf{54.1} & 53.1                                                     & 54.7                                                   & 26.4          & 42.6                                                     \\
$\tau$-norm    & 56.2          & 56.0          & 79.7                                                     & 77.9                                                   & 36.5          & 52.0          & 54.3                                                     & 52.3                                                   & 28.1          & 43.3                                                     \\
cRT            & 56.1          & 55.5          & 81.2                                                     & 75.1                                                   & 36.8          & 53.8          & 50.1                                                     & 52.3                                                   & \textbf{28.6} & 43.5                                                     \\
RIDE           & 56.2          & 77.1          & 80.5                                                     & 79.8                                                   & 35.4          & 45.9          & 54.5                                                     & 55.9                                                   & \textbf{28.6} & 44.6                                                     \\ \hline
TLC(2 experts) & 55.6          & \textbf{83.8} & 85.8                                            & 84.2                                                   & 36.8          & \textbf{54.1} & 52.9                                                     & 55.9                                                   & 27.9          & 44.6                                                     \\
TLC(3 experts) & \textbf{56.9} & 74.9          & \textbf{87.0}                                                     & \textbf{86.2}                                          & 36.3          & 53.4          & 52.9                                                     & 53.7                                                   & 28.1          & 43.9                                                     \\
TLC(4 experts) & 56.5          & 80.5          & 82.8                                                     & 84.7                                                   & \textbf{37.3} & \textbf{54.1} & \textbf{54.6}                                            & \textbf{56.5}                                          & \textbf{28.6} & \textbf{44.7}                                            \\ \hline
\end{tabular}
\end{table*}
\noindent\textbf{Tail \& OOD detection (Q2).}
We evaluate the performances on tail detection and OOD detection in terms of AUC scores~\cite{mcclish1989analyzing}. For tail detection, we label the tail classes as positive and the others as negative. For OOD detection, we jointly use the in-distribution and OOD samples by labeling the in-distribution as negative and the OOD as positive. We use the MCP~\cite{hendrycks2016baseline} (maximal value in the softmax distribution) to quantify the uncertainty for the compared methods. The evaluation results are listed in Table.~\ref{tab:ood}. Our proposed TLC outperforms the compared methods, and is especially better at identifying OOD samples in large image datasets (i.e., ImageNet-open and Places-open).

\begin{table*}[ht]\small%
\renewcommand{\arraystretch}{0.84}
\centering
\caption{Performance comparison on failure prediction (in percentage).}
\label{tab:fp}
\begin{tabular}{c|c|cccc|cccc|cccc}
\hline
\multirow{2}{*}{Dataset}      & \multirow{2}{*}{Method} & \multicolumn{4}{c|}{AUC $\uparrow$}                           & \multicolumn{4}{c|}{FPR-95 $\downarrow$}                      & \multicolumn{4}{c}{ECE $\downarrow$}                          \\ \cline{3-14} 
                              &                         & All           & Head          & Med           & Tail          & All           & Head          & Med           & Tail          & All           & Head          & Med           & Tail          \\ \hline\hline
\multirow{9}{*}{CIFAR-10-LT}  & Focal Loss              & 75.7          & 80.5          & 75.6          & 85.1          & 79.8          & 72.5          & 80.9          & 80.3          & 20.1          & 11.7          & 19.7          & 33.3          \\
                              & OLTR                    & \textbf{83.9} & 79.6          & 83.9          & \textbf{85.9} & 79.6          & 72.9          & 82.4          & 80.7          & 18.8          & 11.2          & 19.6          & 33.2          \\
                              & LDAM-DRW                & 83.1          & 79.6          & 86.3          & 85.2          & 69.3          & 71.7          & 72.9          & 62.1          & 18.9          & 12.4          & 22.0          & 24.5          \\
                              & $\tau$-norm             & 83.8          & 79.5          & 85.2          & 83.5          & 67.9          & 71.2          & 71.9          & \textbf{59.7} & 17.8          & 12.0          & 20.7          & 22.3          \\
                              & cRT                     & 83.7          & 79.8          & 84.1          & 85.3          & 67.4          & 71.2          & 70.3          & 62.1          & 18.4          & 11.5          & 19.8          & 21.3          \\
                              & RIDE                    & 82.9          & 81.7          & 85.4          & 84.2          & 68.7          & 71.9          & 70.5          & 62.3          & 15.9          & \textbf{9.8}  & 17.9          & 22.4          \\ \cline{2-14} 
                              & TLC(2 experts)          & 83.5          & \textbf{84.3} & 85.6          & 83.4          & 68.9          & 66.7          & 68.0          & 71.2          & 12.8          & 10.6          & 13.1          & \textbf{15.8} \\
                              & TLC(3 experts)          & 83.7          & 83.0          & 87.1          & 83.6          & 68.0          & 68.7          & 60.8          & 72.7          & 13.1          & 11.3          & 12.4          & 16.8          \\
                              & TLC(4 experts)          & \textbf{83.9} & 84.0          & \textbf{87.8} & 83.8          & \textbf{65.7} & \textbf{66.3} & \textbf{58.2} & 68.2          & \textbf{12.5} & 11.4          & \textbf{11.3} & 15.9          \\ \hline\hline
\multirow{9}{*}{CIFAR-100-LT} & Focal Loss              & 73.3          & 83.7          & 72.9          & 53.3          & 78.9          & 66.4          & 81.2          & 89.5          & 24.2          & 16.0          & 22.3          & 35.0          \\
                              & OLTR                    & 73.5          & 85.6          & 79.2          & 56.3          & 79.5          & 69.5          & 79.6          & 90.4          & 23.6          & 16.2          & 22.2          & 34.5          \\
                              & LDAM-DRW                & 72.7          & \textbf{85.7} & 75.5          & 55.6          & 81.8          & 68.9          & 76.7          & 92.1          & 30.9          & 18.7          & 30.8          & 43.6          \\
                              & $\tau$-norm             & 73.9          & 85.5          & 75.1          & 54.8          & 78.9          & 66.0          & 83.0          & 89.3          & 29.8          & 17.2          & 29.7          & 42.5          \\
                              & cRT                     & 74.1          & 83.4          & \textbf{79.7} & 53.5          & 78.6          & 64.5          & 78.9          & \textbf{89.2} & 30.2          & 19.8          & 28.9          & 43.8          \\
                              & RIDE                    & 76.3          & 85.5          & 79.5          & \textbf{60.0} & 79.5          & 66.2          & 80.1          & 89.7          & 24.1          & \textbf{14.5} & 23.8          & 34.3          \\ \cline{2-14} 
                              & TLC(2 experts)          & \textbf{77.9} & \textbf{85.7} & 78.7          & \textbf{60.0} & \textbf{78.3} & \textbf{64.3} & 77.7          & 91.2          & 23.2          & 27.8          & 23.0          & \textbf{22.4} \\
                              & TLC(3 experts)          & 76.9          & 84.5          & 78.8          & 57.5          & 79.8          & 67.1          & \textbf{76.5} & 90.6          & 22.8          & 24.8          & 21.9          & 24.6          \\
                              & TLC(4 experts)          & 76.7          & 85.3          & 77.7          & 58.6          & 80.5          & 66.8          & 80.6          & 89.8          & \textbf{21.2} & 21.7          & \textbf{20.6} & 25.6          \\ \hline\hline
\multirow{9}{*}{ImageNet-LT}  & Focal Loss              & 65.4          & 73.7          & 62.8          & 43.4          & 83.9          & 68.4          & 86.2          & 94.5          & 35.3          & 28.3          & 33.2          & 45.1          \\
                              & OLTR                    & 66.0          & 72.8          & 67.0          & 42.3          & 82.9          & 67.1          & 85.0          & 96.3          & 34.8          & 25.6          & 32.6          & 42.3          \\
                              & LDAM-DRW                & 66.8          & \textbf{76.8} & 67.5          & 47.6          & 82.8          & 70.9          & 81.7          & 96.1          & 35.1          & 28.6          & 40.7          & 50.4          \\
                              & $\tau$-norm             & 66.1          & 71.6          & 59.7          & 48.5          & 82.9          & 68.4          & 83.1          & 94.7          & 33.0          & 27.8          & 38.1          & 50.4          \\
                              & cRT                     & 66.4          & 70.2          & 63.3          & 49.2          & \textbf{82.6} & \textbf{65.1} & 83.2          & 95.3          & 32.5          & 27.2          & 37.4          & 48.8          \\
                              & RIDE                    & 66.2          & 75.8          & 70.3          & 47.1          & 84.5          & 68.2          & 85.1          & 94.7          & 31.9          & 24.5          & 35.7          & 42.4          \\ \cline{2-14} 
                              & TLC(2 experts)          & 66.7          & 75.2          & \textbf{71.2} & 48.2          & 84.5          & 65.9          & \textbf{80.5} & 94.5          & 32.8          & 26.3          & \textbf{30.7} & 40.3          \\
                              & TLC(3 experts)          & 66.7          & 75.7          & 68.3          & 48.8          & 84.6          & 67.0          & 80.8          & 97.2          & \textbf{31.3} & 24.7          & 31.6          & \textbf{39.6} \\
                              & TLC(4 experts)          & \textbf{67.2} & 76.2          & 68.7          & \textbf{49.4} & \textbf{82.6} & 65.2          & 81.5          & \textbf{94.3} & 31.9          & \textbf{23.8} & 30.8          & 40.1          \\ \hline
\end{tabular}
\end{table*}

\noindent\textbf{Failure prediction (Q2).} We evaluate the performances on failure prediction in terms of AUC~\cite{mcclish1989analyzing}, the FPR at 95\% TPR (FPR-95)~\cite{liang2017principled} and the Expected Calibration Error (ECE)~\cite{naeini2015obtaining} respectively for the head, medium and tail classes. We also use the MCP~\cite{hendrycks2016baseline} to quantify uncertainty for the compared methods. The evaluation results are listed in Table.~\ref{tab:fp}. Our TLC outperforms the compared methods, and performs much better especially in terms of ECE.

\subsection{Qualitative Evaluation\label{ssec:visual}}
\begin{figure}[ht]
    \centering
    \includegraphics[width=7cm]{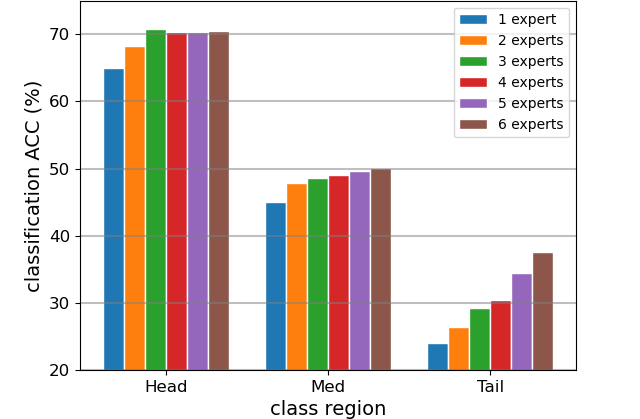}
    \caption{Classification accuracy on CIFAR-100-LT for the head, medium and tail classes with different number of experts.}
    \label{fig:reduce}
\end{figure}
\noindent\textbf{Number of experts (Q3).}
We visualize the accuracy on three class regions (head, medium and tail) on CIFAR-100-LT with ascending numbers of experts in Fig.~\ref{fig:reduce}. We find that using more experts is beneficial to the tail classes, but do not have significant effect for the head classes. This observation justifies our motivation that assigning the same number of experts is redundant for easy samples.

\begin{figure}[ht]
    \centering
    \includegraphics[width=7cm]{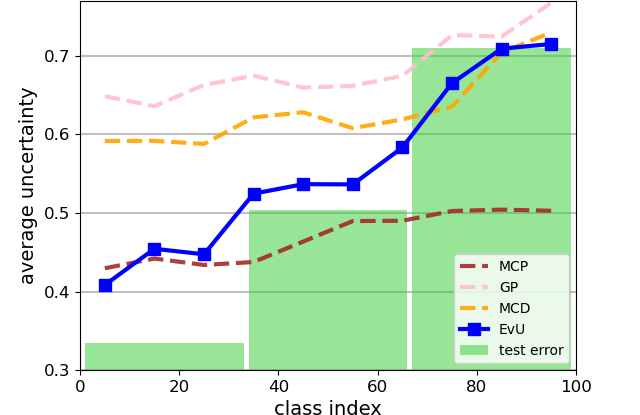}
    \caption{Average uncertainty on CIFAR-100-LT from different uncertainty estimation algorithms (also with the test errors of 3 class regions as benchmarks).}
    \label{fig:uncertainty}
\end{figure}
\noindent\textbf{Uncertainty for each class (Q4).}
We visualize the uncertainty of each class of the CIFAR-100-LT test data with various uncertainty estimation algorithms in Fig.~\ref{fig:uncertainty} (using 3 experts for the evidence-based uncertainty, EvU for short). We compute the averaged uncertainties of every 10 classes. Among the compared uncertainties, EvU is the most consistent with the real test errors. Therefore, it is easy to distinguish the head, medium and tail classes with EvU (practically, low uncertainty indicates the head classes and high uncertainty indicates the tail classes).

\begin{figure}[ht]
    \centering
    \includegraphics[width=8cm]{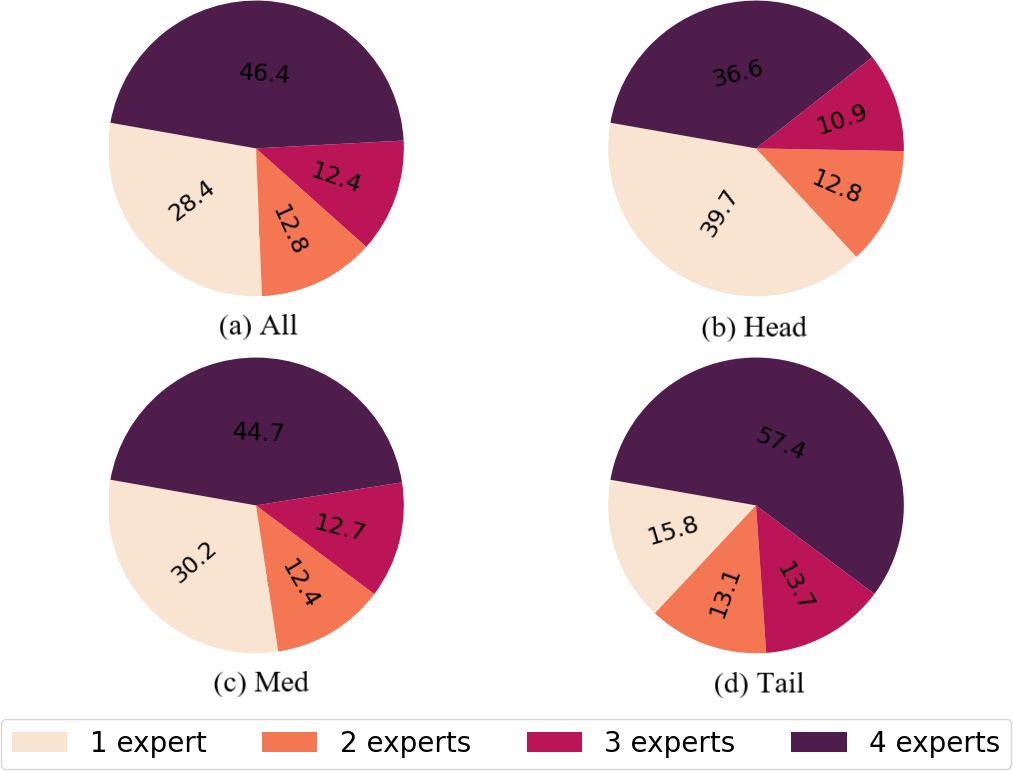}
    \caption{Visualization of expert engagement on CIFAR-100-LT (the percentage of samples using specific number of experts are marked on the pie charts).}
    \label{fig:engage}
\end{figure}
\noindent\textbf{Expert engagement (Q5).}
We visualize the percentage of samples using diverse number of experts respectively on the head, medium, tail and all classes on CIFAR-100-LT in Fig.~\ref{fig:engage}. We set the maximal number of experts as 4 and the threshold $\tau=0.54$ according to the hyperparameter settings of quantitative evaluations. The samples using 4 experts dominate the tail classes and the samples using 1 expert dominate the head classes. Overall, the hard samples are assigned with more experts to learn the patterns, which is consistent with our motivation in Sec.~\ref{ssec:learn}.

\subsection{Ablation Study\label{ssec:ablation}}
\begin{table}[ht]\small%
\setlength\tabcolsep{2.3pt}
\renewcommand{\arraystretch}{0.84}
\centering
\caption{Ablation study on different combination of objective components of the proposed TLC with CIFAR-100-LT dataset.}
\label{tab:component}
\begin{tabular}{ccc|ccc}
\hline
$\mathcal{L}$     & $\mathcal{L}_{kl}$     & $w$          & \begin{tabular}[c]{@{}c@{}}ACC\\ classification\end{tabular} & \begin{tabular}[c]{@{}c@{}}AUC\\ OOD detection\end{tabular} & \begin{tabular}[c]{@{}c@{}}AUC\\ failure prediction\end{tabular} \\ \hline
$\checkmark$ &              &              & 48.9                                                         & 45.8                                                        & 63.7                                                             \\
$\checkmark$ & $\checkmark$ &              & \textbf{49.1}                                                & \textbf{53.2}                                               & \textbf{76.2}                                                    \\
$\checkmark$ &              & $\checkmark$ & 48.7                                                         & 42.4                                                        & 61.6                                                             \\ \hline\hline
$\checkmark$ & $\checkmark$ & $\checkmark$ & 49.0                                                         & 53.4                                                        & 76.9                                                             \\ \hline
\end{tabular}
\end{table}
\noindent\textbf{Effectiveness of components.}
We compare different combinations of the components ($\mathcal{L}, \mathcal{L}_{kl}$ and the prefix weight $w$) on CIFAR-100-LT in terms of classification, OOD detection and failure prediction (all using 3 experts). The results are listed in Table.~\ref{tab:component}, where we also include the results of full objective for reference. It is easy to conclude: \romannumeral1) adding $\mathcal{L}_{kl}$ is beneficial to obtaining more reliable uncertainty (comparing line 2 against line 1 and line 3 against line 4), and \romannumeral2) dynamically reducing engaged experts (with $w$) does not significantly affect the performance on the three tasks (comparing line 2 with line 4).

\begin{table}[ht]\small%
\renewcommand{\arraystretch}{0.84}
\centering
\caption{Ablation study on uncertainty, comparing different uncertainties on CIFAR-100-LT dataset (in percentage).}
\label{tab:as_uncertainty}
\begin{tabular}{cc|cccc}
\hline
\multicolumn{2}{c|}{Method}                                      & MCP           & Entropy       & MCS           & EvU           \\ \hline
\multicolumn{1}{c|}{\multirow{4}{*}{AUC $\uparrow$}}      & All  & 74.3          & 75.1          & 76.4          & \textbf{77.9} \\
\multicolumn{1}{c|}{}                                     & Head & 83.9          & 84.6          & 84.4          & \textbf{85.7} \\
\multicolumn{1}{c|}{}                                     & Med  & 79.2          & 78.9          & \textbf{80.9} & 78.7          \\
\multicolumn{1}{c|}{}                                     & Tail & 57.0          & 57.6          & 58.2          & \textbf{60.0} \\ \hline
\multicolumn{1}{c|}{\multirow{4}{*}{FPR-95 $\downarrow$}} & All  & 79.5          & 79.4          & 80.9          & \textbf{78.3} \\
\multicolumn{1}{c|}{}                                     & Head & 66.0          & 65.6          & 65.4          & \textbf{64.3} \\
\multicolumn{1}{c|}{}                                     & Med  & 76.1          & 76.8          & \textbf{75.7} & 77.7          \\
\multicolumn{1}{c|}{}                                     & Tail & 89.7          & \textbf{89.1} & 90.1          & 91.2          \\ \hline
\multicolumn{1}{c|}{\multirow{4}{*}{ECE $\downarrow$}}    & All  & 24.1          & 25.0          & 23.9          & \textbf{23.2} \\
\multicolumn{1}{c|}{}                                     & Head & \textbf{14.5} & 18.5          & 19.7          & 27.8          \\
\multicolumn{1}{c|}{}                                     & Med  & 23.8          & 24.3          & 23.1          & \textbf{23.0} \\
\multicolumn{1}{c|}{}                                     & Tail & 34.3          & 34.7          & 32.6          & \textbf{22.4} \\ \hline
\end{tabular}
\end{table}

\noindent\textbf{Comparison of uncertainties (Q2).}
We compare various uncertainty estimation algorithms on failure prediction. We use 1 expert as the backbone model and compute MCP~\cite{hendrycks2016baseline}, GP~\cite{damianou2013deep}, MCD~\cite{gal2016dropout} and the EvU on CIFAR-100-LT dataset. According to the results in Table.~\ref{tab:as_uncertainty}, the EvU outperforms the other algorithms on all tasks especially on the tail classes, which validates that the EvU is more reliable than other compared uncertainty estimation algorithms.

\section{Conclusion\label{sec:conclusion}}
In this paper, we propose the Trustworthy Long-tailed Classification (TLC), which estimates evidence and uncertainty in a multi-expert framework. The estimated evidence and uncertainty of each expert are combined under the Dempster-Shafer Evidence Theory (DST). The TLC can dynamically reduce the number of engaged experts for easy samples, which ensures efficiency while preserving promising performances. We evaluate the TLC on multiple tasks with diverse metrics, where it outperforms existing methods and is trustworthy with reliable uncertainty.

\section*{Acknowledgements}
This work was partly supported by the National Natural Science Foundation of China (61976151, 61732011), the National Key Research and Development Program of China under Grant 2019YFB2101900, and the A*STAR AI3 HTPO Seed Fund (C211118012). We gratefully acknowledge the support of MindSpore, CANN (Compute Architecture for Neural Networks) and Ascend AI Processor used for this research.

%%%%%%%%% REFERENCES
{\small
\bibliographystyle{ieee_fullname}
\bibliography{main}
}

\end{document}